\documentclass[conference]{IEEEtran}
\IEEEoverridecommandlockouts
\usepackage{amsmath,amssymb,amsfonts}
\usepackage{algorithmic}
\usepackage{graphicx}
\usepackage{textcomp}
\usepackage{xcolor}
\usepackage{balance}
\def\BibTeX{{\rm B\kern-.05em{\sc i\kern-.025em b}\kern-.08em
    T\kern-.1667em\lower.7ex\hbox{E}\kern-.125emX}}
\begin{document}

\title{Solving Math Word Problems Using Estimation Verification and Equation Generation\\
}

\author{\IEEEauthorblockN{Mitchell Piehl}
\IEEEauthorblockA{\textit{Computer Science} \\
\textit{University of Iowa}\\
Iowa City, USA \\
mpiehl@uiowa.edu }
\and
\IEEEauthorblockN{Dillon Wilson}
\IEEEauthorblockA{\textit{Computer Science} \\
\textit{University of Colorado}\\
Colorado Springs, USA \\
dwilso21@uccs.edu}
\and
\IEEEauthorblockN{Ananya Kalita}
\IEEEauthorblockA{\textit{Foster School of Business} \\
\textit{University of Washington}\\
Seattle, Washington \\
ananyak@uw.edu}
\and
\IEEEauthorblockN{Jugal Kalita}
\IEEEauthorblockA{\textit{Computer Science} \\
\textit{University of Colorado}\\
Colorado Springs, USA \\
jkalita@uccs.edu}

}
\maketitle

\begin{abstract}
Large Language Models (LLMs) excel at various tasks, including problem-solving and question-answering. However, LLMs often find Math Word Problems (MWPs) challenging because solving them requires a range of reasoning and mathematical abilities with which LLMs seem to struggle. Recent efforts have helped LLMs solve more complex MWPs with improved prompts. This study proposes a novel method that initially prompts an LLM to create equations from a  decomposition of the question, followed by using an external symbolic equation solver to produce an answer. To ensure the accuracy of the obtained answer, inspired by an established recommendation of math teachers, the LLM is instructed to solve the MWP a second time, but this time with the objective of estimating the correct answer instead of solving it exactly. The estimation is then compared to the generated answer to verify. If verification fails, an iterative rectification process is employed to ensure the correct answer is eventually found. This approach achieves new state-of-the-art results on datasets used by prior published research on numeric and algebraic MWPs, improving the previous best results by nearly two percent on average. In addition, the approach obtains satisfactory results on trigonometric MWPs, a task not previously attempted to the authors' best knowledge. This study also introduces two new datasets, SVAMPClean and Trig300, to further advance the testing of LLMs' reasoning abilities.\footnote{The code, results, and datasets can be found at: https://github.com/mitchellpiehl/EVoSS}
\end{abstract}

\begin{IEEEkeywords}
Math word problems, Large language models, Answer estimation, Trigonometric word problems. 
\end{IEEEkeywords}

\section{Introduction}
Solving MWPs has been an area of interest in natural language processing since Daniel Bobrow in 1964 \cite{Bobrow} developed a computing system to understand natural language and solve algebraic word problems. 
In recent years, researchers have found many different ways of solving MWPs, including using Seq2Seq methods~\cite{Wang2017}, Seq2Tree methods ~\cite{Wang2019}, \cite{Xie}, Graph2Tree methods~\cite{Zhang} or using transformers to directly translate from text to algebraic equations~\cite{Griffith2020}, \cite{Griffith2021} and many more. 
Recently, there has been a noticed surge in the use of LLMs to solve MWPs to achieve high accuracy~\cite{Brown}, which has been difficult to obtain despite the perceived simplicity of MWPs. 
LLMs often get confused when solving MWPs unless assisted in reasoning steps via appropriate prompts \cite{Shi}. 
To address this challenge, researchers have aided plain LLMs with Chain-of-Thought prompting, which allows the LLM to solve an MWP as a multi-step reasoning task ~\cite{Wei}, \cite{Kojima}. 
Among the various state-of-the-art prompt engineering techniques, Progressive Rectification Prompting combined with Chain-of-Thought~\cite{Wu} stands out. 
This method allows an LLM to rectify its answer by generating responses until the answer passes verification. Another notable technique, BRIDGE, erases the interrogative parts of an MWP, decomposes the rest of the MWP into simpler statements, translates them into equations, and finally solves to get the answer~\cite{Wang}.

\begin{figure*}[t]
    \centering
    \includegraphics[width=\textwidth]{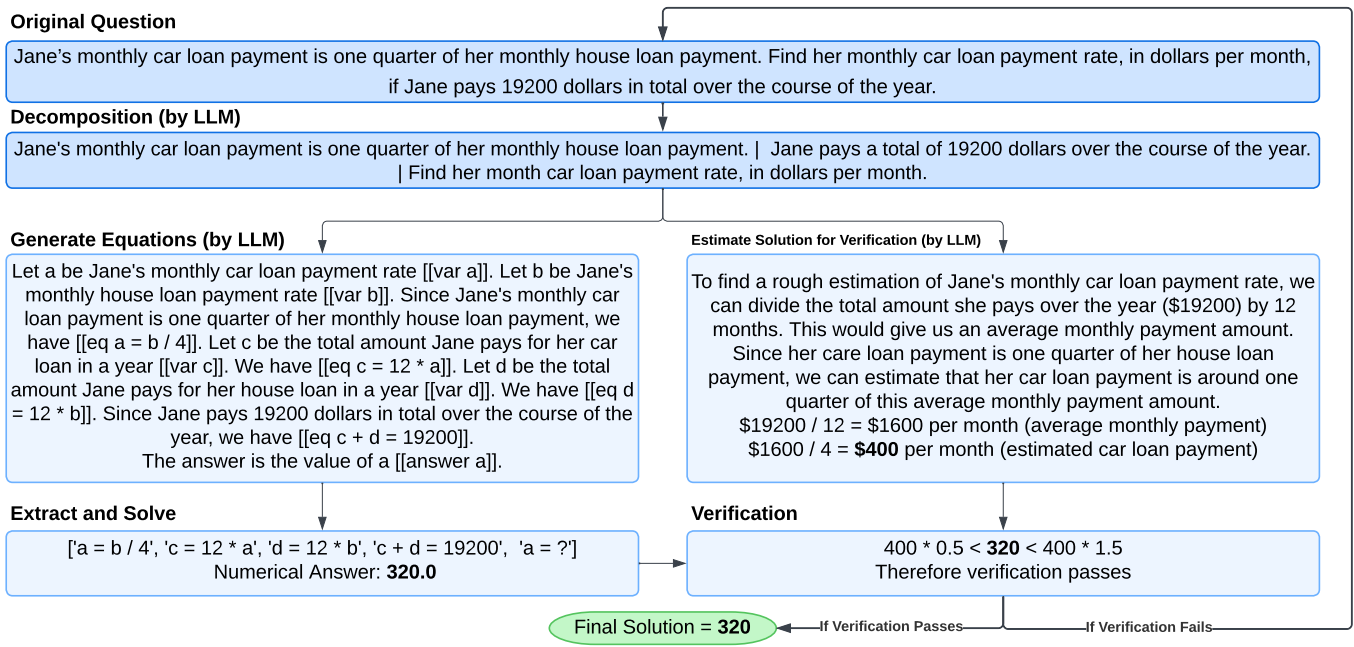}
    \caption{Simplified diagram example of the EVoSS method from the Algebra dataset}
    \label{fig:False}
\end{figure*}

Current methods for solving MWPs, in spite of being aided by designed prompts, sometimes solve problems incorrectly due to various reasons. 
This paper proposes a novel approach to solve MWPs called Equation Verification of Symbolic Solvers (EVoSS). Figure 1 illustrates EVoSS with an example from the Algebra dataset \cite{He}. The method starts by instructing an LLM to decompose the problem statement and generate algebraic equations. Then, a symbolic solver, external to the LLM, is used to solve the problem precisely. 
The proposed method does not commit the generated answer as final right away, but uses a verification technique that asks the LLM to estimate the answer instead of solving exactly. By comparing the rough estimate with the precise answer, the approach increases the chances of discovering mistakes during equation generation. The introduction of the answer-estimation step during the solving process is inspired by a commonsense practice among teachers of elementary math, supported by publications, which recommend checking the obtained answer against an estimate when the work is finished ~\cite{National}, \cite{reed1984estimating}, \cite{kilpatrick1967analyzing}, \cite{faulk1962well}. 
There are many YouTube videos\footnote{This video at the well-respected site Khan Academy discusses a two-step estimation of answers to word problems: https://www.khanacademy.org/math/cc-third-grade-math/arithmetic-patterns-and-problem-solving/estimation-word-problems/v/2-step-estimation-word-problems. } and websites geared toward elementary school children that discuss how to estimate an answer to math word problems and check the reasonableness of the estimate. 
The approach is as follows. 

The first step involves asking the LLM to decompose the question into more straightforward statements and removing unnecessary textual details that may hinder correct translation into equations. Once the equations have been generated, a symbolic solver attempts to solve them.
Once the symbolic solver has computed a numerical value, the LLM is presented with the question again. However, the model is now asked to estimate the correct answer instead of solving it precisely. This estimate serves as a point of comparison with the numerical value derived earlier. If the estimation is aligned with the answer, the verification process confirms that the correct answer has been found. 
If verification fails due to the estimate significantly diverging from the answer obtained by the symbolic solver, a rectification process finds an answer again. During this process, the LLM is presented with the original question and the estimation as a hint to guide the LLM to the correct answer.

This paper makes several key contributions. First, the paper introduces a novel MWP answer verification scheme that uses a prompting-based method, assisted by a symbolic solver to verify answers generated by MWPs. This method achieves new state-of-the-art results. Secondly, the paper creates a new trigonometry dataset that is utilized to stress-test the symbolic solver and estimation verification. Finally, this paper finds inadmissible errors in a commonly used dataset and fixes them so that future solvers will be tested fairly. 

The rest of the paper starts by discussing related works on math word problem-solving. The following sections present the problem statement, and introduces the approach that uses answer estimation to verify MWP answers. Finally, the paper elaborates upon the experiments performed, the results obtained, and the new datasets that are published for use by future researchers.

\section{Related Work}
A number of recent methods use LLMs to solve MWPs. Here are a variety of state-of-the-art methods that use diverse methodologies.

\subsection{CoT Prompting Methods}
Chain-of-thought prompting methods ask LLMs to generate reasoning paths while solving MWPs to guide the process ~\cite{Wei}. 
The reasoning path helps the LLM solve the MWP because 
the process of explicitly generating the steps in the path and articulating them in natural language, often causes the LLM to correct any mistakes it may have committed when solving a problem
\cite{Tan}. 
Two CoT prompting methods have been proposed recently: zero-shot prompting ~\cite{Kojima} and few-shot prompting ~\cite{Wei}. Zero-shot CoT prompting instructs the LLM to find the answer without any examples. In contrast, few-shot prompting gives the LLM a few examples illustrating the instructions before asking the LLM to solve. Auto-CoT automatically generates reasoning chains by sampling questions to minimize manual effort and provide more diverse examples ~\cite{Zhang2023}. Self-consistency (SC) has improved CoT prompting by identifying and selecting the most frequently obtained answer across diverse reasoning paths ~\cite{Wang2023self}. 

\subsection{Progressive Rectification Prompting Methods}
CoT and self-consistency tend to repeat the same mistakes. Progressive Rectification Prompting (PRP) aims to address these errors by verifying an answer through a process where one numerical value is removed from the MWP, and the large language model is tasked with identifying the missing value ~\cite{Wu}. If the LLM does not correctly find the missing value, the answer can be rectified using a hint provided to the LLM, then verified again. 

\subsection{Decomposition Methods}
A few methods decompose the question before finding the answer to assist the LLM. For example, the BRIDGE technique achieves high solution accuracy by using a 4-step process that proceeds as follows: erase, decompose, translate, and answer ~\cite{Wang}. Another method, Decomposed Prompting, solves complex tasks using a modular decomposition structure to optimize specific sub-tasks ~\cite{Khot}.

\subsection{Methods Using External Solvers}
A common way to solve MWPs is to use an external equation solver after the LLM generates the necessary equations, using Program-of-Thoughts prompting (PoT). PoT uses LLMs to generate program statements and then delegates the statements to a program interpreter ~\cite{Chen}. Another example, the Program-aided Language Model or PaL, generates programs as the intermediate reasoning steps and then uses a runtime environment such as a Python Interpreter to avoid mathematical errors by the LLM~\cite{Gao}. Finally, another state-of-the-art technique, declarative prompting, uses an LLM to formalize word problem variables and equations. This process mitigates arithmetic errors using an external symbolic solver that solves the equations without error ~\cite{He}. This method beats PaL by around 20 percent on their new dataset consisting of more complex word problems that we will also use to test our model, called Algebra. 

\section{Problem Statement}
\label{sec:problemStatement}
Despite many successes using various methodologies, there is still room for improving the effectiveness of LLMs in solving  math word problems (MWPs). Current methods often suffer due to an increased chance of errors in problems requiring multiple equations and a lack of appropriate validation. In addition, current approaches to MWP solving have not ventured out of simple numerical and algebra problems. Our approach, Estimation Verification of Symbolic Solver Results (EVoSS), aims to develop a model that achieves higher accuracy in solving MWPs, particularly for MWPs that require multiple equations to solve or have large non-rounded numerical values by adding flexibility to the verification process. 

\section{Approach}
This study introduces a novel estimation-based verification technique that validates the answer obtained while solving LLM-generated equations from the LLM-decomposed question using an external symbolic solver. Specifically, this method takes a given question, asks the LLM to decompose it into  components that are easier to translate into equations and turn the decomposition into equations, finds an answer from the equations 
and finally finds an estimation from the LLM and compares that to the prior generated answer to verify and rectify if needed. 

\subsection{Pre-processing and Decomposition}
In pre-processing and decomposition, the LLM is asked to separate the question sentence-wise into two parts: a statement part and an actual question part. The question is usually the entire last sentence or is contained in the last sentence, but need not always be. The statements and the asking part are presented to the LLM. A two-shot prompting approach that provides examples of wanted outputs increases accuracy and directs the LLM toward the desired outcome.

\subsection{Equation Generation}
The LLM is prompted to create equations from the decomposition in a specific format because the decomposition is often more straightforward to turn into equations than the original question. We use a set of simple rules to guide the LLM while generating the equations.\footnote{An appendix of the exact rules provided during equation generation can be found at: https://github.com/mitchellpiehl/EVoSS/tree/main/FinalSolver} Three-shot prompting, which gives three examples of the decomposition and generated equations, is used here to increase the accuracy of equation generation. 

\subsection{Solving Equations}
The equations generated by the LLM are extracted
and sent to the symbolic solver. The generated system of equations may be unsolvable due to issues such as being over-determined or having no unique solutions. To combat these errors, several checks are in place to identify any errors returned by the symbolic solver. If errors are found, the LLM is prompted to generate new equations with a helper statement reminding the LLM of the broken rule. For example, if the last sentence of the original question does not give the goal, we will provide the hint: ``Make sure your response ends with 'The answer is the value of x [[answer x]].', where x is the goal variable." If the LLM has been prompted several times to improve the set of equations, the last set of equations are sent to the LLM to solve, and the process moves on. We set the value of the maximum attempts to generate equations to 3 for all tests. Once the symbolic solver has generated a numerical answer that is not an error, the answer is compared to the estimation to be verified, as discussed in the next Subsection \ref{subsec:estimationVerification}.
Our approach  uses a symbolic solver to 
solve the generated equations. 
Any error in equation generation by the LLM will result in an incorrect answer. Thus, incorporating an additional step of answer verification using estimation improves the possibility of finding the correct result.

\subsection{Estimation Verification}
\label{subsec:estimationVerification}
When generating equations, two outcomes are possible. Either the LLM is entirely accurate with no mistakes, resulting in the correct answer, or it introduces one or more errors in the equations, leading to significantly incorrect answers. For this reason, estimation verification is likely to work well. 
Inspired by  common sense and supported by a publication from the National Council of Teachers of Mathematics ~\cite{National}, during estimation verification, the LLM is given the question decomposition and prompted to find a rough estimate of the final answer. 
Once the estimation is obtained, it is compared against the answer generated by the symbolic solver. The comparison involves checking whether the estimation is within $\alpha\%$ of the generated answer. Ablation studies has shown that $\alpha$ varies between different datasets ranging from 40 to 50. Table 4 displays the accuracy of various parameter values on the three datasets. If the estimation closely matches the generated answer, the verification passes, indicating confidence in the equations' accuracy. If the estimation is not close to the generated answer, the verification fails, and rectification ensues. 

\subsection{Rectification}
When verification fails, the question will be prompted to the LLM another time, but this time, the estimation will be given as a hint to guide the LLM. To gain more utility from the hint, the LLM will not be asked to generate equations but instead simply find an answer. Once the answer is generated, if it has been generated in previous steps, our method declares that answer as the final solution. The same answer must be generated at least 2 out of 3 times to be accepted as correct.

\begin{table*}[t]
    \caption{Accuracy Comparison of ten methods on four datasets. The highest accuracy per dataset is bold, and the second highest is underlined. The average of the four datasets is also shown.}
    \label{tab:methods3}
    \centering
    \begin{tabular}{l l l l l | l}
        \hline
        {Method} & {GSM8K} & {SVAMP} & {Algebra} & {SVAMPClean} & {Average} \\
        \hline
        EVoSS (Ours) & \underline{82.2} & \textbf{89.4} & \textbf{92.8} & \textbf{90.5} & \textbf{88.73} \\
        I3C & \textbf{84.3} & 85.8 & \underline{89.2} & \underline{88.4} & \underline{86.93} \\
        BRIDGE & 77.2 & 82.3 & 82.0 & {--} & {80.50} \\
        PoT & 76.3 & \underline{88.2} & 64.0 & {--} & {76.17} \\
        Auto-CoT & 78.8 & 80.9 & 53.6 & {--} & {71.10} \\
        PAL & 79.5 & 77.8 & {--} & {--} & {78.65} \\
        Zero-Shot-CoT & 78.6 & 82.7 & 82.8 & 84.3 & {82.10} \\
        PS & 75.9 & 77.8 & {--} & {--} & {76.85} \\
        Manual-CoT & 76.4 & 82.7 & 77.9 & 85 & {80.50} \\
        Direct & 77.8 & 80.6 & 65.3 & 83.3 & {76.75} \\
        \hline
    \end{tabular}
\end{table*}
\section{Experiments}
We tested our method using the three most relevant datasets, plus two new datasets, and compared the results with the most recent and significant methods while ensuring comparable implementations.\footnote{The code, results, and datasets can be found at:  https://github.com/mitchellpiehl/EVoSS}

\begin{figure}
    \centering    \includegraphics[scale=0.8]{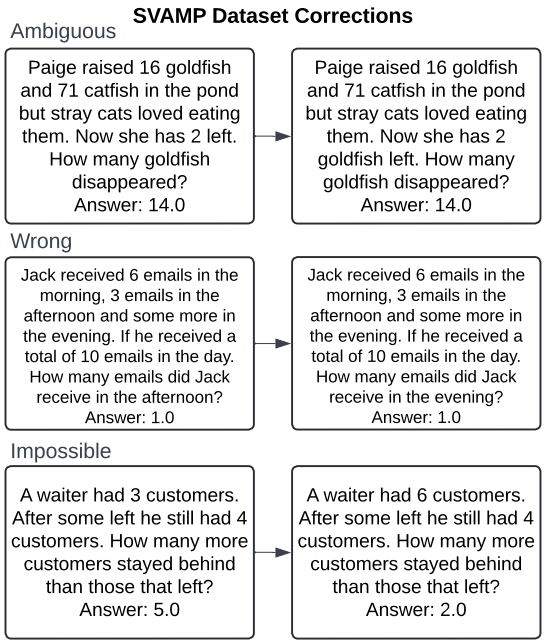}
    \caption{Three examples of corrections made to the SVAMP dataset to create the new SVAMPClean Dataset}
    \label{fig:corrections}
\end{figure}

\subsection{Datasets}
This work has used three different datasets to test the proposed method. GSM8k~\cite{Cobbe}, SVAMP~\cite{Patel}, and Algebra ~\cite{He}. GSM8k is a dataset of 1,319 grade school MWPs commonly used to test MWP solvers. The SVAMP dataset has 1000 math questions similar to the GSM8K dataset, which requires no more than 2 basic arithmetic operations to solve. Algebra is a more recent dataset consisting of 222 algebra-based questions from textbooks made for middle school students instead of elementary students. In addition to these datasets, while analyzing the SVAMP dataset, it was found that many questions were faulty because the questions were simply wrong, ambiguous, or impossible to occur in real life, so we fixed these errors and ran additional testing on the modified dataset, which we call SVAMPClean. This new SVAMPClean dataset is the same as before; however, 50 questions have been improved to make them correct, more solvable, or more specific. The difficulty level of the questions remained the same as that of the original SVAMP data set. Figure 2 shows an example of each type of error in the original dataset and how it was fixed. 

\begin{figure}
    \centering    \includegraphics[scale=0.65]{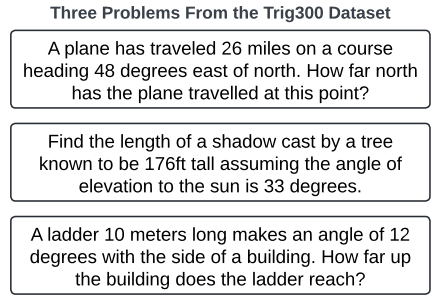}
    \caption{Three examples of questions in the new Trig300 dataset}
    \label{fig:Trig300}
\end{figure}

As MWP solvers and LLMs become more advanced, more challenging datasets will be required to test the reasoning ability and growth of different models and methods, so we created a new dataset called Trig300. This dataset contains 300 trigonometry-based questions that require the solver to go beyond addition, subtraction, multiplication, and division to solve the problem correctly. All 300 questions require the use of one of the trigonometric operations. This dataset also tests how estimation verification works on questions beyond addition, subtraction, multiplication, and division. To create this dataset, each question was hand-crafted to mimic various questions found in trigonometry textbooks and then manipulated to ask for different parts based on the information provided. This ensures the dataset tests the solver's diverse solving skills. For example, the question, "If a treadmill is 48 inches long and is set to an incline of 10 degrees. How much higher is the front of the treadmill than the back?" will be changed to also ask for the angle of incline of the treadmill given the length and height increase, and it will also ask for the length of the treadmill given the incline and height increase. All answers are rounded to the nearest thousandth if necessary. Examples of problems from the Trig300 dataset are shown in Figure 3.

\subsection{Baselines}
The results are compared with nine current state-of-the-art methods: Auto Chain of Thought ~\cite{Zhang2023}, BRIDGE ~\cite{Wang}, Direct~\cite{Kojima}, Identify and Ignore Irrelevant Conditions (I3C) ~\cite{Wu2024III}, Manual Chain of Thought (CoT) ~\cite{Wei}, Plan-and-Solve (PS) ~\cite{Wang2023}, Program-Aided Learning (PAL) ~\cite{Gao}, Program of Thought (PoT) ~\cite{Chen}, and Zero-Shot Chain of Thought~\cite{Kojima}. We will also compare our estimation verification technique to self-consistency (SC) ~\cite{Wang2023self}.

\begin{table*}
    \small
        \caption{Case study of various answers generated by the listed methods. The question is underlined, and the responses generated by our method are in bold.}
    \label{tab:methods1}
    \centering
    \begin{tabular}{p{1in} p{4in}}
        \hline
        \underline{Question:} & {A monthly pass for light rail is 1.2 times cheaper than a bundle with 20 day passes. Find the price of the monthly pass if the bundle costs \$10 more than the monthly pass. Answer: 50} \\
        \hline
        Direct & The answer is 30. \textbf{Incorrect}\\
        \hline
        I3C & Step 1: Let x be the price of the monthly pass. Step 2: Since the bundle with 20 day passes costs \$10 more than the monthly pass, the price of the bundle is x + \$10. Step 3: According to the given condition, the monthly pass is 1.2 times cheaper than the bundle with 20 day passes, so we can write the equation: x = 1.2(x + \$10). Step 4: Now we can solve for x: x = 1.2x + \$12, 0.2x = \$12, x = \$60. Step 5: Therefore, the price of the monthly pass is \$60. \textbf{Incorrect} \\
        \hline
        Zero-shot CoT & 1. Let the price of the monthly pass be M. 2. Since the bundle costs \$10 more than the monthly pass, the price of the bundle is M + 10. 3. According to the statement, the monthly pass is 1.2 times cheaper than the bundle with 20 day passes. This means: M = 1.2(M + 10). 4. Solving the equation: M = 1.2M + 12, 0.2M = 12, M = 60. 5. Therefore, the price of the monthly pass is \$60. \textbf{Incorrect} \\
        \hline
        \textbf{Equation Generation (Ours)} & Let a be the price of the monthly pass [[var a]]. Let b be the price of the bundle with 20 day passes [[var b]]. We have [[eq b = a + 10]]. Since the bundle costs \$10 more than the monthly pass, we have [[eq b = 1.2 * a]]. The answer is the value of a [[answer a]]. \textbf{Correct}\\
        \hline
        \textbf{Estimation (Ours)} & To find a rough estimation, we can assume the monthly pass costs \$x. According to the first statement, the bundle with 20 day passes costs 1.2 times more, so it would be 1.2x. And according to the second statement, the bundle costs \$10 more than the monthly pass, so 1.2x = x + 10. Solving for x: 1.2x = x + 10, 0.2x = 10, x = 50. Therefore, the price of the monthly pass is roughly \$50. \textbf{Correct}\\
        \hline
    \end{tabular}
\end{table*}
\normalsize

\begin{table}
    \caption{Ablation study results comparing the use of the Symbolic Solver with Estimation Verification (EVoSS) to the use of the Symbolic Solver with decomposition (SS) and the use of estimation verification technique without the symbolic solver.}
    \label{tab:SSEV}
    \centering
    \begin{tabular}{l l l l}
        \hline
        Method & {GSM8K} & {SVAMP} & {Algebra} \\
        \hline
        EVoSS & \textbf{82.2} & \textbf{89.4} & \textbf{92.8} \\
        SS & 69.9 & 80.8 & 79.7 \\
        EV & 74.1 & {85.0} & {76.1} \\
        \hline
    \end{tabular}
\end{table}

\subsection{Implementation}
This method uses the gpt-3.5-turbo API as the backend large language model\footnote{The public API is available at https://openai.com/api/}. The gpt-3.5-turbo API is widely accessible and is an advanced large language model easily available to researchers for minimal cost. It has also been used to test many current state-of-the-art methods. We set the temperature to 0.4 and the max tokens to 256.

A few-shot reasoning approach is used for question decomposition and equation generation, which provides examples of questions and responses to help guide the LLM. Similar to the Declarative ~\cite{He} and BRIDGE ~\cite{Wang} prompting methods, the equations generated are solved using SymPy ~\cite{Meurer}, a Python library for symbolic mathematics. 

\subsection{Metric}
We evaluate our answers using the exact match metric. This means that we count our answers as correct if the answer matches exactly. However, due to round-off errors and the Trig300 dataset answers being rounded to the nearest thousandth, we count our answers as correct if the generated answer is within 0.001 of the correct answer. 

\subsection{Results}
Table 1 reports the accuracy of our method across the specified data sets. Our novel approach achieves state-of-the-art performance on the Algebra, SVAMP, and SVAMPClean datasets, achieving accuracies of 92.8, 89.4, and 90.5 percent, respectively. These results beat its nearest competitor by around 3 percent in all three datasets. Our state-of-the-art results achieve an accuracy of 88.73 percent across the four datasets, beating the next-best average by almost two percent.

\subsection{SVAMPClean Dataset}
After the inadmissible errors were corrected in the SVAMP dataset, all methods tested received a higher accuracy with SVAMPClean than the original SVAMP dataset. The improvements from SVAMP to SVAMPClean are consistently between one and three percent, coherent with the five percent of questions being modified. 

\subsection{Analysis}
In addition to the presented results, an ablation study was performed on the datasets with and without estimation verification and a symbolic solver. These findings are summarized in Table 3. The results show improvements from eight to thirteen percent using estimation verification across all three datasets tested, demonstrating how our novel estimation verification can drastically improve equation generation solvers. 

\begin{table}[ht]
    \caption{Accuracies when changing the percentage of how close the estimation needs to be away from the answer to pass verification. Temperature was set to 0.7 for all tests.}
    \label{tab:Alphas}
    \centering
    \begin{tabular}{l l l l}
        \hline
        {$\alpha$} & {GSM8K} & {SVAMP} & {Algebra} \\
        \hline
        40 & 78.0 & 85.3 & \textbf{88.3} \\
        50 & \textbf{80.0} & \textbf{87.9} & 87.8 \\
        60 & 79.0 & 83.9 & 87.3 \\
        \hline
    \end{tabular}
\end{table}

Testing was also done to see how results are affected when the accuracy needed to pass verification changes. Table 4 presents these results. The results show that the optimal $\alpha$ value while using the Algebra dataset was 40, while the optimal $\alpha$ value using the GSM8K and SVAMP datasets was 50.

\begin{table}[ht]
    \caption{Accuracies when changing the temperature of the LLM}
    \centering
    \begin{tabular}{l l l l}
        \hline
        Temperature & {GSM8K} & {SVAMP} & {Algebra} \\
        \hline
        0.7 & 80 & 87.9 & 88.3 \\
        0.4 & \textbf{82.2} & \textbf{89.4} & \textbf{92.8}\\
        0.1 & 80.8 & 89.2 & 88.7 \\
        \hline
    \end{tabular}
    \label{tab:Temperature}
\end{table}

In addition to the tests above, we have obtained the MWP solver's accuracies using different LLM temperatures. The results are shown in Table 5. This method is most accurate when the LLM temperature is set to 0.4 in all three datasets tested.

\begin{table}[ht]
    \caption{Comparing our method of Estimation Verification to using the symbolic solver with self-consistency}
    \label{tab:EVvsSC}
    \centering
    \begin{tabular}{l l l l}
        \hline
        Method & {GSM8K} & {SVAMP} & {Algebra} \\
        \hline
        EVoSS& \textbf{82.2} & \textbf{89.4} & \textbf{92.8} \\
        SCoSS & 75.1 & {86.6} & 90.0 \\
        \hline
    \end{tabular}
\end{table}

\begin{table*}[t]
     \caption{Case study and example of a problem being solved using our methods from an example of the new Trig300 dataset.}
    \label{tab:methods2}
    \centering
    \begin{tabular}{p{1in} p{4in}}
        \hline
        \textbf{Question:} & {Pacing off 150 ft. on flat ground from the base of a tree, a forester measures the angle of elevation to the top of the tree as 40◦. What is the height of the tree? Answer: 125.865} \\
        \hline
        {Equation Generation } & Let a be the distance from the base of the tree [[var a]]. We have [[eq a = 150]]. Let b be the height of the tree [[var b]]. Let c be the angle of elevation to the top of the tree in radians [[var c]]. We have [[eq c = 40 * pi / 180]]. Since the angle of elevation to the top of the tree forms a right triangle with the distance and the height, we have [[eq tan(c) = b / a]]. The answer is the value of b [[answer b]]. \textbf{Correct}\\
        \hline
        {Estimation} & To find a rough estimation of the height of the tree, we can use the trigonometric function tangent. Tangent(angle) = opposite/adjacent, Tangent(40 degrees) = height /150 ft, height = 150 ft * tangent(40 degrees), height \~ 150 ft * 0.84 \~ 125 ft. Therefore, the rough estimation of the height of the tree is approximately 125 feet. \textbf{Correct}\\
        \hline
    \end{tabular}
\end{table*}
\normalsize

Finally, we compared our estimation verification technique to a current state-of-the-art technique, self-consistency (SC). Self-consistency is the act of repeatably finding the answer to the same question \textit{M} times and selecting the final answer as the most frequent answer of all the generated answers ~\cite{Wang2023self}. For our experiments, we replaced our estimation verification with self-consistency by removing our estimation verification technique and repeatedly generating equations and solving using the symbolic solver. We set \textit{M} to 5 for our Self-Consistency of Symbolic Solver (SCoSS) experiments. Our results show that using estimation to check answers generated achieves two to seven percent higher accuracy than using self-consistency while using a symbolic solver. Estimation verification is superior to self-consistency due to the tendency of the LLM to get confused in the same ways when solving the question repeatedly. Using estimation to verify answers generated creates diverse solving paths and simplifies the problem, leading to more accurate results. The results are shown in Table 6.

\subsection{Case Study}
Table 2 presents a case study of various answers generated by a real example from the Algebra dataset. As questions become layered with more statements, standard solving techniques, such as the ones shown, cannot handle the multiple layers well and get the answer incorrect. However, solving the question by creating equations to be solved by a symbolic solver, or simplifying the question by estimating, both of which are used in our method, leads to a correct answer generated. 

\begin{table}[ht]
    \caption{Accuracy percentage results of 4 different methods tested on the Trig300 dataset}
    \label{tab:TrigResults}
    \centering
    \begin{tabular}{p{3cm} p{2cm}}
        \hline
        Method & {Trig300} \\
        \hline
        EVoSS & \textbf{65.5} \\
        I3C & 17 \\
        Zero Shot-CoT & 10.7 \\
        Manual-CoT & 10 \\
        Direct & 4 \\
        \hline
    \end{tabular}
\end{table}

\subsection{Trig300 Dataset}
We ran testing of our new Trig300 dataset on our method and three other baseline methods. Our method achieved 65.5 percent accuracy, compared to 4 - 17 percent accuracy on the other baseline methods. The obtained accuracy ranging between 4 and 66 percent across the four methods compared to the other datasets ranging from 53 to 93 percent, shows how this new dataset is significantly more challenging for LLMs, thus providing an improved way of testing the reasoning abilities of LLMs as they continue to get more advanced. In addition, our method beating the other baseline methods by 50 to 60 percent, shows how our method is superior to other baseline models, especially when solving more complex math problems. Additionally, this shows how estimation verification can be successfully implemented on other problems beyond the four basic operators: addition, subtraction, multiplication, and division. Table 8 shows our results, and Table 7 shows an example of a question in the new dataset and a case study of the question being solved using equation generation and estimation.

\section{Potential Risks and Downfalls}
The increasing use of large language models (LLMs) presents both opportunities and risks within the educational system that need to be recognized. On one hand, improved reasoning in LLMs provide access to higher-quality information and problem solving tools. However, on the other hand, over-reliance on LLMs discourages critical thinking abilities and the development of problem-solving skills. 

A second potential limitation concerns the longevity of the proposed method. As LLMs continue to evolve and improve, they may naturally become better at solving MWPs without the help of outside methods such as the proposed method. For instance, GPT-4 solved MWPs from the GSM8K dataset with nearly 35 percent higher accuracy than GPT-3.5. ~\cite{OpenAI}. The rapid progression of LLMs may lead to augmented approaches becoming less impactful, especially as newer models demonstrate stronger baselines. It is important to note that as performance improves, datasets of more complex word problems from various domains will have to be developed to ensure progress in evaluating the mathematical and reasoning abilities of the LLMs. 

\section{Conclusion}
This study presents a novel estimation verification technique that verifies solutions obtained from an improved symbolic solver called EVoSS. This method leverages question decomposition to facilitate more accurate equation generation so that a symbolic solver can be used. The answer is then verified using an estimation verification technique that prompts the model to find an estimate and compares it to the obtained answer. The estimation verification technique improves results by eight to thirteen percent, helping this method solve math word problems with state-of-the-art accuracy. Additionally, our work fixes a popular dataset, SVAMP, by removing inadmissible errors and ambiguities. We call the new dataset SVAMPClean. Finally, this paper creates a new, more complex trigonometry dataset to prepare for the advancement of various methods and models and to test if the estimation verification works beyond the simple MWPs based on the four operators: addition, subtraction, multiplication, and division. 

\section{Acknowledgements}
All work herein reported is supported by the National Science Foundation under Grant No. 2349452. Any opinion, finding, or conclusion in this study is that of the authors and does not necessarily reflect the views of the National Science Foundation.

\balance
\bibliography{acl2016}
\bibliographystyle{unsrt}
\end{document}